\documentclass[10pt,twocolumn,letterpaper]{article}

\usepackage[pagenumbers]{cvpr} 

\definecolor{cvprblue}{rgb}{0.21,0.49,0.74}
\usepackage[pagebackref,breaklinks,colorlinks,allcolors=cvprblue]{hyperref}
\usepackage{graphicx}
\usepackage{amsmath}
\usepackage{amssymb}
\usepackage{booktabs}
\usepackage{mmstyles}
\usepackage{multirow}
\usepackage{makecell}
\usepackage{siunitx}
\usepackage{algorithm}
\usepackage{listings}
\usepackage[accsupp]{axessibility}  
\usepackage{etoolbox}
\makeatletter
\AfterEndEnvironment{algorithm}{\let\@algcomment\relax}
\AtEndEnvironment{algorithm}{\kern2pt\hrule\relax\vskip3pt\@algcomment}
\let\@algcomment\relax
\newcommand\algcomment[1]{\def\@algcomment{\footnotesize#1}}
\renewcommand\fs@ruled{\def\@fs@cfont{\bfseries}\let\@fs@capt\floatc@ruled
  \def\@fs@pre{\hrule height.8pt depth0pt \kern2pt}%
  \def\@fs@post{}%
  \def\@fs@mid{\kern2pt\hrule\kern2pt}%
  \let\@fs@iftopcapt\iftrue}
\makeatother

\graphicspath{{figures/}}
\DeclareGraphicsExtensions{.pdf}

\def\confName{CVPR}
\def\confYear{2025}

\title{Test-Time Visual In-Context Tuning}

\author{
Jiahao Xie$^{1,2}$, Alessio Tonioni$^{3}$, Nathalie Rauschmayr$^{3}$, Federico Tombari$^{3}$, Bernt Schiele$^{1,2}$\\
$^{1}$Max Planck Institute for Informatics, SIC \quad 
$^{2}$VIA Research Center \quad 
$^{3}$Google\\
\tt\small \{jxie, schiele\}@mpi-inf.mpg.de \quad \{alessiot, rauschmayr, tombari\}@google.com
}

\begin{document}

\twocolumn[{%
\renewcommand\twocolumn[1][]{#1}%
\maketitle
\begin{center}
    \centering 
    \includegraphics[width=\linewidth]{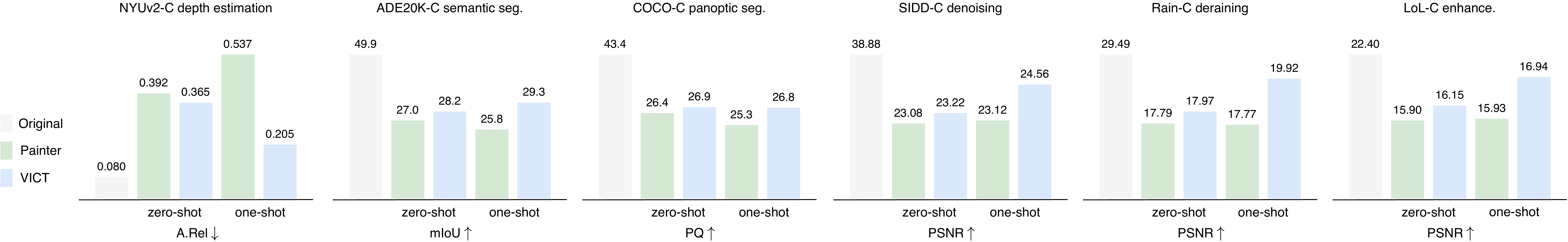}
    \captionof{figure}{
    \textbf{Test-time visual in-context tuning (VICT) on six representative vision tasks under distribution shifts.} We benchmark the robustness of VICL with 15 common corruptions adopted in~\cite{hendrycks2019benchmarking,michaelis2019benchmarking}, and report the averaged performance across all corruptions. Existing VICL models like Painter exhibit poor generalization capability to unseen new domains when the task prompts come from the training distribution (\ie, \emph{zero-shot}). Performances are even worse when given task prompts from the test distribution (\ie, \emph{one-shot}). By performing VICT at test time, we can significantly improve Painter in both zero-shot and one-shot manners.
    }
    \label{fig:teaser}
\end{center}%
}]


\begin{abstract}
Visual in-context learning (VICL), as a new paradigm in computer vision, allows the model to rapidly adapt to various tasks with only a handful of prompts and examples. While effective, the existing VICL paradigm exhibits poor generalizability under distribution shifts. In this work, we propose test-time \textbf{V}isual \textbf{I}n-\textbf{C}ontext \textbf{T}uning (\textbf{VICT}), a method that can adapt VICL models on the fly with a single test sample. Specifically, we flip the role between the task prompts and the test sample and use a cycle consistency loss to reconstruct the original task prompt output. Our key insight is that a model should be aware of a new test distribution if it can successfully recover the original task prompts. Extensive experiments on six representative vision tasks ranging from high-level visual understanding to low-level image processing, with 15 common corruptions, demonstrate that our VICT can improve the generalizability of VICL to unseen new domains. In addition, we show the potential of applying VICT for unseen tasks at test time. Code: \url{https://github.com/Jiahao000/VICT}.
\end{abstract}

\section{Introduction}
\label{sec:intro}

Following the success of in-context learning (ICL)~\cite{brown2020language,hao2022language,alayrac2022flamingo} in natural language processing (NLP), visual in-context learning (VICL)~\cite{bar2022visual,wang2023images} has shown promising performance in developing generalist models for vision tasks.
Inspired by prompting in NLP that defines tasks using language sequences as a general interface, existing VICL works~\cite{bar2022visual,wang2023images} use images themselves as a natural interface for general-purpose visual perception. They formulate VICL as an image inpainting task, \ie, given an input-output example describing a specific task (\ie, the prompt) as well as an input image, they are assembled in a grid and the problem is casted as inpainting the missing part of the grid (\ie, the prediction) consistently with the given task prompts. This enables adapting a pre-trained vision model to various downstream tasks with only a handful of prompts and examples.

By default, current VICL models are frozen during deployment. However, their performance might suffer in real-world deployment as the test distribution usually changes and deviates from the training one. This naturally raises a question: what is the generalizability of VICL models under distribution shifts? In this work, we investigate this aspect focusing on controllable shifts due to image corruptions. We observe that the existing VICL paradigm like Painter~\cite{wang2023images} exhibits poor generalizability to unseen new domains when the task prompt comes from the training distribution, as shown in Figure~\ref{fig:teaser}. Surprisingly, performances are even worse when given input-output task prompts belonging to the same distribution as the test input. 

Driven by this observation, we propose to rely on test-time visual in-context tuning (VICT) to adapt VICL models on the fly to unseen distributions using only the given test sample.
The motivation is that each test input offers a hint about the test distribution. Thus, we modify the VICL model at test time to make full use of this hint by setting up a one-sample learning problem. 
Specifically, given input-output task prompts and the test input, we first use the VICL model to inpaint the output image. We then flip the role between the task prompts and the test sample, \ie, we treat the predicted test output as the prompt to the model and reconstruct the original output of task prompts.
This allows us to tune the parameters of the whole model in a self-supervised manner that can be applied to arbitrary tasks. 
Our key insight is that a model should be aware of a new test distribution if it can successfully recover the original task prompts conditioned on its in-context inference. Such a cycle consistency supervision signal naturally exists in the context of VICL without requiring any additional training data or annotations, thus making it an appealing self-supervisory task for test-time visual in-context training.

We explore VICT in two settings: (\romannumeral1) zero-shot setting, where the task prompts are from the training distribution (\ie, clean images), and (\romannumeral2) one-shot setting, where the task prompts are from the test distribution (\ie, corrupted images).
Without loss of generality, we consider Painter~\cite{wang2023images} as our VICL model, for its simplicity in design and its wide applicability.
As shown in Figure~\ref{fig:teaser}, our simple method leads to substantial improvements across 15 common corruptions~\cite{hendrycks2019benchmarking,michaelis2019benchmarking} on six representative vision tasks ranging from high-level visual understanding to low-level image processing, including depth estimation on NYUv2~\cite{silberman2012indoor}, semantic segmentation on ADE20K~\cite{zhou2017scene}, panoptic segmentation on COCO~\cite{lin2014microsoft}, image denoising on SIDD~\cite{abdelhamed2018high}, image deraining on the merged deraining dataset~\cite{zamir2020learning}, and low-light image enhancement on LoL~\cite{wei2018deep}.

Our main contributions are summarized as follows:

\textbf{1)} We propose a new cycle consistency task for test-time visual in-context tuning. To the best of our knowledge, we are the first to perform test-time training for VICL.

\textbf{2)} We contribute the first study on the generalizability of VICL under distribution shifts. We observe that the existing VICL paradigm exhibits poor generalizability to unseen new domains. Such a phenomenon can hardly be recovered even given input-output task prompts with the same distribution as the test input. 

\textbf{3)} We conduct extensive experiments on six representative vision tasks across 15 common corruptions. VICT significantly improves Painter in both zero-shot and one-shot manners. Our zero-shot or one-shot VICT can even outperform Painter trained with more few-shot corrupted examples. We also explore the potential of applying VICT for unseen tasks at test time, further demonstrating its promise.


\section{Related Work}
\label{sec:work}

\noindent\textbf{In-context learning.}
\emph{In-context learning} has been extensively explored in the NLP literature after the introduction of large language models. Seminal works in the field like GPT-3~\cite{brown2020language} go as far as claiming that language models are indeed few-shot learners. All recent language models like~\cite{touvron2023llama,touvron2023llama2,dubey2024llama3} have the ability to adapt their behavior based on few-shot examples provided in their context window or directly to follow simple user instructions~\cite{chung2024scaling}. 
Similar concepts have been extended to multimodal models with seminal works like Flamingo~\cite{alayrac2022flamingo} showing few-shot capabilities by providing examples as interleaved text and images in the context window. Some of these capabilities are now available for extremely large commercial multimodal large language models~\cite{team2023gemini,team2024gemini,openai2023gpt}.

Apart from using languages as the general interface, a recent line of work introduces purely \emph{visual in-context learning} (VICL) by training vision generalist models that can perform arbitrary visual tasks following one or few visual examples provided at inference time. Two representative works developed in parallel are: (\romannumeral1) MAE-VQGAN~\cite{bar2022visual}, which trains a variant of MAE~\cite{he2022masked} on a dataset of figures extracted from academic papers, and (\romannumeral2) Painter~\cite{wang2023images}, which uses a similar idea but trains its model on a set of standard academic benchmark datasets. More recently, this strategy has been expanded in LVM~\cite{bai2024sequential}, where even more visual datasets are collected to train an autoregressive generative model.
We base our work on Painter~\cite{wang2023images} due to its simplicity in design and its wide applicability. 
It is the one that provides both the available code and model weights, and has already been extensively evaluated across several standard benchmarks.

\noindent\textbf{Generalization under distribution shifts.}
Machine learning models suffer from performance drops when tested on a data distribution different from the one they are trained on~\cite{vasiljevic2016examining,geirhos2018generalisation,tsipras2019robustness,zhang2019theoretically}. The lower the drop, the more we define a model robust or able to generalize to distribution shifts. Common strategies to increase model robustness include using heavy data augmentations~\cite{zheng2016improving,dodge2017quality,tobin2017domain,hendrycks2019benchmarking} or extremely large training distributions going up to web scale~\cite{radford2021learning,zhai2023sigmoid,naeem2024silc}. Nevertheless, most of the experiments are done in a classification setting. To the best of our knowledge, there is no previous work exploring the robustness of VICL models. We are the first to counteract performance drops in VICL under distribution shifts.

\begin{figure*}[t]
    \centering
    \includegraphics[width=.99\linewidth]{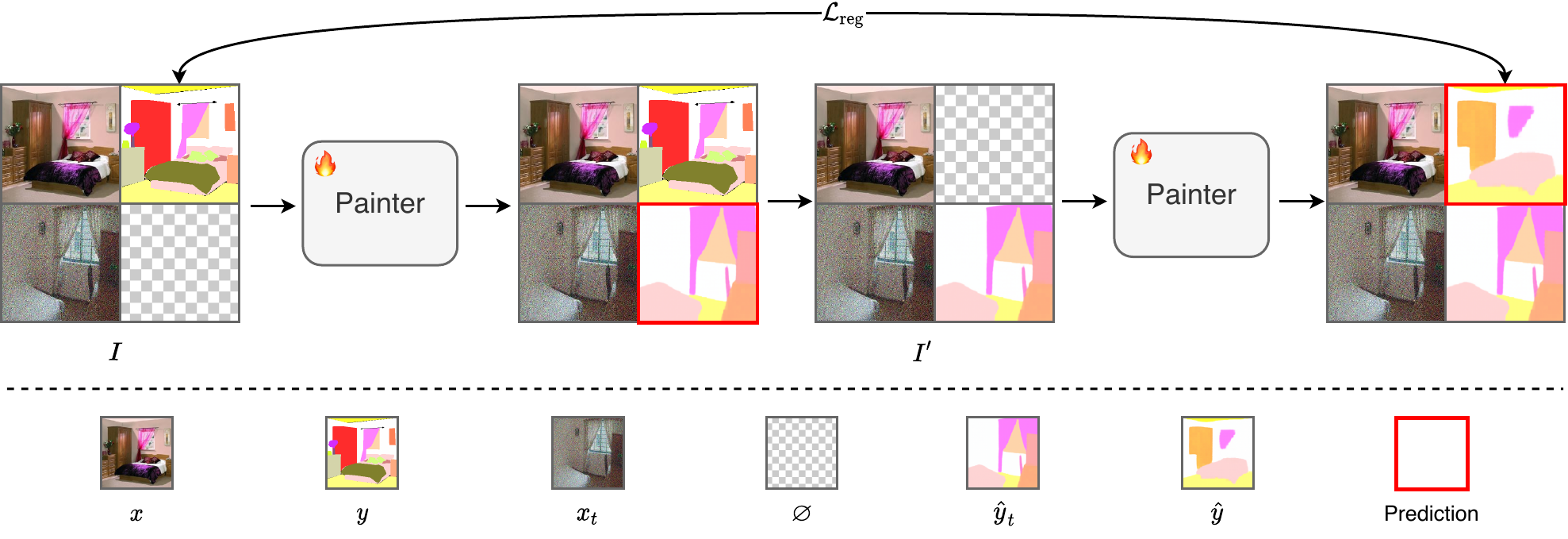}
    \vspace{-5pt}
    \caption{\textbf{Overview of our VICT pipeline.} Given a pair of task prompts $\left(x,y\right)$ and a test input image $x_t$, we first construct a four-cell grid-like image canvas $I=\left(x,y,x_t,\varnothing\right)$, with an empty cell at the bottom right. We then feed $I$ into the VICL model (\eg, Painter) to predict the test output $\hat{y}_t$. Afterward, we flip the role between input-output task prompts and input-output test samples, \ie, we provide the predicted $\hat{y}_t$ as the \emph{prompt} to the model, recreating a new four-cell grid-like image canvas $I^{\prime}=\left(x,\varnothing,x_t,\hat{y}_t\right)$, with an empty cell at the top right. The new $I^{\prime}$ is fed into the same model to predict the task prompt output $\hat{y}$. We finally optimize the model by minimizing the distance between $\hat{y}$ and $y$ via a standard regression loss.}
    \label{fig:framework}
    \vspace{-5pt}
\end{figure*}

\noindent\textbf{Test-time training.}
An alternative way of counteracting performance drops due to distribution shifts is test-time training~\cite{sun2020test,liu2021ttt++,wang2021tent,bartler2022mt3,chen2022contrastive,gandelsman2022test,tsai2023convolutional}. This accounts to unfreezing the model at test time and fine-tuning it on the target distribution (or a single sample from it) through self-supervision. Early works propose this paradigm using self-supervised pretext tasks (\eg, rotation prediction~\cite{gidaris2018unsupervised}) to explicitly improve generalization under image corruptions~\cite{sun2020test} or improve in specific tasks where self-supervised losses can be clearly defined like depth estimation~\cite{tonioni2019real,poggi2021continual,li2023test}, reinforcement learning~\cite{hansen2021self,mutti2024test},  tracking~\cite{fu2021learning,segu2023darth}, and NLP~\cite{banerjee2021self,ye2022robust}.
More recently, the field has realized that MAE~\cite{he2022masked} provides a very strong self-supervised pretext task and has therefore been employed in the context of test-time training~\cite{gandelsman2022test}.
Finally, some recent works~\cite{shu2022test,feng2023diverse,ma2023swapprompt,karmanov2024efficient,zanella2024test} also employ the test-time training paradigm for vision-language models to improve their generalization.
To the best of our knowledge, ours is the first work to explore test-time training applied to VICL models, where previous self-supervised pretext tasks~\cite{doersch2015unsupervised,noroozi2016unsupervised,gidaris2018unsupervised,zhan2020online,chen2020simple,xie2022delving,he2022masked,xie2023masked,li2023correlational} cannot be employed due to their constraints on the single-image context.


\section{Methodology}
\label{sec:method}

Our visual in-context tuning (VICT) is a simple yet effective test-time training approach to adapt visual in-context learning (VICL) models on the fly.
In this section, we first introduce some preliminaries on VICL in Section~\ref{subsec:vicl}. We then detail our VICT pipeline in Section~\ref{subsec:vict}.

\subsection{Visual In-Context Learning}
\label{subsec:vicl}

In-context learning is a new paradigm originating from large language models such as GPT-3~\cite{brown2020language} in NLP. Unlike traditional learning paradigms, in-context learning formulates different NLP tasks as text completion tasks and makes predictions conditioned on one or many support examples provided to the model in the context window. Extending this paradigm to tasks requiring a dense visual output in the form of images is nontrivial and has not been addressed in the literature for a long time. Recently, \cite{bar2022visual,wang2023images} formulate VICL as an image inpainting task by combining images and labels into a grid-like new image and using masked image modeling for pre-training, granting the models with the in-context learning ability.

Formally, let $S=\{\left(x_i,y_i\right)\}_{i=1}^{N}$ denote the set of support input-output examples (\aka, task prompts) where $x$ is an image and $y$ is its visual label (\eg, a segmentation mask).\footnote{Following previous works, we set $N=1$ in practice.} Given $S$ and a new test image $x_t$ as input, VICL can be formulated as follows:
\begin{equation}
y_t=f_\theta(S, x_t),
\end{equation}
where $f_\theta\left(*\right)$ is the VICL model parametrized with $\theta$, $y_t$ is the corresponding output label of $x_t$.
Usually, we assume $S$ and $x_t$ are drawn from the same distribution. However, this is rarely the case for real-world deployment. Therefore, in this work, we mainly consider the scenario where a distribution shift like an image corruption occurs for $x_t$.

In the experiment section, we will show how the presence of such distribution shifts significantly degrades the performance of the VICL models.
Moreover, we will additionally show how simple solutions like providing in-context examples from the target (corrupted) distribution would not fix the problem due to the limited generalization ability of the current VICL models.

\subsection{Test-Time Visual In-Context Tuning}
\label{subsec:vict}

In this work, we propose to leverage test-time training~\cite{sun2020test} to counteract the distribution shifts between training and testing data, with the goal of making VICL models more robust. 
We argue that VICL models are particularly amenable to be optimized at test time on arbitrary tasks thanks to the availability of the $N$ shots provided as context.
The core intuition of our method is to use the provided in-context examples to define a self-supervised loss that can be used at test time to train any VICL models on the fly on any task. The overall pipeline of our VICT is illustrated in Figure~\ref{fig:framework}.

\begin{algorithm}[t]
\caption{Pseudocode of VICT in a PyTorch-like style.}
\label{alg:code}
\definecolor{codeblue}{rgb}{0.25,0.5,0.5}
\definecolor{codekw}{rgb}{0.85, 0.18, 0.50}
\lstset{
  backgroundcolor=\color{white},
  basicstyle=\fontsize{8pt}{8pt}\ttfamily\selectfont,
  columns=fullflexible,
  breaklines=true,
  captionpos=b,
  commentstyle=\fontsize{8pt}{8pt}\color{codeblue},
  keywordstyle=\fontsize{8pt}{8pt}\color{codekw},
}
\begin{lstlisting}[language=python]
# f: VICL model (e.g., Painter)
# theta_0: pre-trained model weights
# x: task prompt input
# y: task prompt output
# mask_token: mask token for the empty cell
# steps: number of test-time optimization steps

for x_t in loader: # load a test sample x_t
    f.params = theta_0 # initialize
    # test-time optimization
    for t in range(steps):
        # predict the test sample output
        y_t_pred = f((x, y, x_t, mask_token))
        # predict the task prompt output
        y_pred = f((x, mask_token, x_t, y_t_pred))
        
        # regression loss
        loss = SmoothL1Loss(y_pred, y)

        # update model
        loss.backward()
        update(f.params)

    # inference with the updated model
    y_t_pred = f((x, y, x_t, mask_token))
\end{lstlisting}
\end{algorithm}

Specifically, given a pair of task prompts $\left(x,y\right)$ from $S$ and a test input $x_t$, we construct a grid-like image canvas with four cells, denoted by an ordered quadruple\footnote{The image order is top left, top right, bottom left, and bottom right, respectively.} $I=\left(x,y,x_t,\varnothing\right)$, where $\varnothing$ is an empty cell. We then feed $I$ into the VICL model $f_\theta$ to predict the test output $\hat{y}_t$:
\begin{equation}
\hat{y}_t=f_\theta\left(I\right)=f_\theta\left(\left(x,y,x_t,\varnothing\right)\right).
\end{equation}
Afterward, with the obtained test output, we flip the role between input-output task prompts and input-output test samples, creating a new four-cell grid-like image canvas with the order of $I^{\prime}=\left(x,\varnothing,x_t,\hat{y}_t\right)$.
The new $I^{\prime}$ is fed into $f_\theta$ again to predict the task prompt output $\hat{y}$: 
\begin{equation}
\hat{y}=f_\theta\left(I^{\prime}\right)=f_\theta\left(\left(x,\varnothing,x_t,\hat{y}_t\right)\right).
\end{equation}
We then use a simple regression loss in pixel space to optimize the model by minimizing the distance between $\hat{y}$ and $y$:
\begin{equation}\label{eq:loss}
\theta_{x_t}=\text{arg}\mathop{\text{min}}\limits_{\theta}\mathcal{L}\left(\hat{y},y\right)
\end{equation}
In practice, we follow~\cite{wang2023images} to use smooth-$\ell_1$~\cite{girshick2015fast} loss. 
Using this loss we can optimize the parameters $\theta$ of the model for a small numbers of steps for each $x_t$ and, eventually, make a prediction on $x_t$ as $f_{\theta_{x_t}}\left(\left(x,y,x_t,\varnothing\right)\right)$. 

In our formulation, the gradient-based optimization for Equation~\ref{eq:loss} always starts from the initial pre-trained model weights $\theta_0$ for each test input. When a new test input arrives, we discard $\theta_{x_t}$ and reset the weights to $\theta_0$.
We follow this strategy to not have any assumption on the test inputs and treat them independently at test time, \ie, we do not assume that they come from the same distribution.
Our generic formulation is completely agnostic on the VICL models considered, on the type of tasks being solved, and on the type of in-context samples provided to it.

The pseudo-code of VICT is in Algorithm~\ref{alg:code}.


\section{Experiments}
\label{sec:exp}

\begin{table*}[t]
\centering
\small
\caption{\textbf{System-level comparison on six representative vision tasks across 15 corruptions.} Results are on corruption level 5. We consider two settings: (\romannumeral1) zero-shot setting, where the task prompts are from the training distribution (\ie, clean images), and (\romannumeral2) one-shot setting, where the task prompts are from the test distribution (\ie, corrupted images). ``avg'' denotes the averaged results over 15 corruptions.}
\label{tab:results}
\addtolength{\tabcolsep}{-2pt}
\begin{tabular}{lcccccccccccccccc}
\toprule
method & brigh & cont & defoc & elast & fog & frost & gauss & glass & impul & jpeg & motn & pixel & shot & snow & zoom & avg \\ \hline\midrule
\multicolumn{17}{c}{(a) depth estimation NYUv2-C (A.Rel $\downarrow$)} \\ \midrule
\multicolumn{17}{l}{\emph{zero-shot setting:}} \\
Painter & 0.129 & 0.215 & 0.712 & 0.109 & 0.129 & 0.536 & 0.200 & 0.612 & 0.189 & 0.386 & 0.167 & 0.187 & 0.142 & 1.951 & 0.209 & 0.392 \\
VICT & 0.108 & 0.216 & 0.631 & 0.108 & 0.133 & 0.576 & 0.191 & 0.541 & 0.181 & 0.310 & 0.139 & 0.134 & 0.140 & 1.856 & 0.210 & 0.365 \\ \midrule
\multicolumn{17}{l}{\emph{one-shot setting:}} \\
Painter & 0.126 & 0.285 & 0.743 & 0.109 & 0.132 & 0.853 & 0.901 & 0.622 & 0.901 & 0.392 & 0.174 & 0.194 & 0.440 & 1.964 & 0.212 & 0.537 \\
VICT & 0.097 & 0.193 & 0.180 & 0.107 & 0.128 & 0.227 & 0.241 & 0.210 & 0.278 & 0.159 & 0.121 & 0.114 & 0.150 & 0.662 & 0.214 & 0.205 \\ \hline\midrule
\multicolumn{17}{c}{(b) semantic segmentation ADE20K-C (mIoU $\uparrow$)} \\ \midrule
\multicolumn{17}{l}{\emph{zero-shot setting:}} \\
Painter & 40.4 & 11.7 & 27.9 & 31.2 & 35.0 & 23.3 & 24.8 & 25.6 & 26.1 & 38.6 & 31.8 & 40.4 & 25.9 & 9.4 & 13.1 & 27.0 \\
VICT & 40.9 & 12.7 & 28.2 & 31.7 & 36.4 & 23.7 & 25.5 & 26.3 & 27.4 & 38.8 & 31.9 & 40.9 & 28.0 & 17.3 & 13.6 & 28.2 \\ \midrule
\multicolumn{17}{l}{\emph{one-shot setting:}} \\
Painter & 40.9 & 11.7 & 27.5 & 31.3 & 35.0 & 22.1 & 19.7 & 25.8 & 20.3 & 38.5 & 31.3 & 40.3 & 23.2 & 6.5 & 13.3 & 25.8 \\
VICT & 41.7 & 21.0 & 28.8 & 32.2 & 37.0 & 24.7 & 24.4 & 27.0 & 25.2 & 39.7 & 32.8 & 41.5 & 26.6 & 21.7 & 14.9 & 29.3 \\ \hline\midrule
\multicolumn{17}{c}{(c) panoptic segmentation COCO-C (PQ $\uparrow$)} \\ \midrule
\multicolumn{17}{l}{\emph{zero-shot setting:}} \\
Painter & 38.1 & 15.5 & 25.5 & 30.0 & 33.3 & 26.0 & 24.9 & 23.8 & 25.5 & 31.9 & 27.6 & 34.9 & 26.3 & 19.8 & 12.5 & 26.4 \\
VICT & 38.7 & 15.7 & 25.8 & 30.3 & 33.9 & 26.7 & 25.3 & 24.3 & 25.9 & 32.4 & 27.7 & 35.4 & 26.7 & 21.9 & 12.7 & 26.9 \\ \midrule
\multicolumn{17}{l}{\emph{one-shot setting:}} \\
Painter & 38.1 & 14.2 & 24.8 & 30.0 & 33.1 & 25.2 & 21.5 & 23.7 & 22.0 & 31.5 & 27.5 & 34.6 & 23.8 & 17.2 & 12.1 & 25.3 \\
VICT & 38.6 & 16.9 & 25.4 & 30.4 & 33.8 & 26.6 & 24.0 & 24.6 & 24.5 & 32.3 & 27.4 & 35.0 & 25.2 & 24.9 & 12.4 & 26.8 \\ \hline\midrule
\multicolumn{17}{c}{(d) denoising SIDD-C (PSNR $\uparrow$)} \\ \midrule
\multicolumn{17}{l}{\emph{zero-shot setting:}} \\
Painter & 8.72 & 25.49 & 32.48 & 33.20 & 16.86 & 8.79 & 17.50 & 33.18 & 17.50 & 23.18 & 29.47 & 34.84 & 25.27 & 9.86 & 29.80 & 23.08 \\
VICT & 8.86 & 25.65 & 32.91 & 32.96 & 17.00 & 8.84 & 17.58 & 33.00 & 17.50 & 24.86 & 29.68 & 34.76 & 25.21 & 9.75 & 29.67 & 23.22 \\ \midrule
\multicolumn{17}{l}{\emph{one-shot setting:}} \\
Painter & 9.12 & 25.16 & 31.99 & 33.23 & 16.68 & 9.24 & 17.99 & 33.14 & 18.33 & 22.73 & 29.30 & 34.87 & 24.69 & 10.50 & 29.76 & 23.12 \\
VICT & 13.94 & 25.32 & 32.41 & 34.45 & 16.45 & 11.15 & 20.13 & 32.96 & 19.93 & 26.80 & 30.11 & 35.01 & 24.93 & 14.90 & 29.95 & 24.56 \\ \hline\midrule
\multicolumn{17}{c}{(e) deraining Rain-C (PSNR $\uparrow$)} \\ \midrule
\multicolumn{17}{l}{\emph{zero-shot setting:}} \\
Painter & 10.74 & 14.15 & 19.54 & 20.10 & 13.91 & 12.49 & 19.85 & 19.80 & 20.16 & 24.86 & 18.52 & 22.81 & 20.11 & 11.89 & 17.93 & 17.79 \\
VICT & 10.89 & 14.24 & 19.58 & 20.31 & 13.94 & 12.62 & 20.42 & 19.83 & 20.67 & 24.97 & 18.56 & 23.05 & 20.47 & 12.05 & 17.98 & 17.97 \\ \midrule
\multicolumn{17}{l}{\emph{one-shot setting:}} \\
Painter & 11.04 & 13.95 & 19.61 & 20.15 & 13.82 & 12.87 & 19.51 & 19.87 & 19.58 & 24.85 & 18.62 & 22.88 & 19.68 & 12.16 & 17.95 & 17.77 \\
VICT & 17.38 & 14.69 & 20.36 & 21.44 & 14.67 & 17.69 & 22.33 & 20.48 & 22.24 & 25.31 & 19.05 & 23.92 & 21.37 & 19.53 & 18.28 & 19.92 \\ \hline\midrule
\multicolumn{17}{c}{(f) low-light enhancement LoL-C (PSNR $\uparrow$)} \\ \midrule
\multicolumn{17}{l}{\emph{zero-shot setting:}} \\
Painter & 16.26 & 13.43 & 18.80 & 19.83 & 11.24 & 11.62 & 12.50 & 18.82 & 12.91 & 15.87 & 18.01 & 21.12 & 17.75 & 13.77 & 16.60 & 15.90 \\
VICT & 16.39 & 13.57 & 19.24 & 19.84 & 11.37 & 11.64 & 12.51 & 19.23 & 13.00 & 16.00 & 18.11 & 21.20 & 19.16 & 13.92 & 17.06 & 16.15 \\ \midrule
\multicolumn{17}{l}{\emph{one-shot setting:}} \\
\multicolumn{1}{l}{Painter} & 16.74 & 13.50 & 18.85 & 19.83 & 11.02 & 11.76 & 13.14 & 18.95 & 13.74 & 15.84 & 18.15 & 21.12 & 16.54 & 13.19 & 16.59 & 15.93 \\
VICT & 18.19 & 15.03 & 18.99 & 19.88 & 11.76 & 12.01 & 14.95 & 19.22 & 14.99 & 16.12 & 18.23 & 21.09 & 19.84 & 16.96 & 16.91 & 16.94 \\ \bottomrule
\end{tabular}
\end{table*}

\begin{figure*}[t]
    \centering
    \includegraphics[width=.99\linewidth]{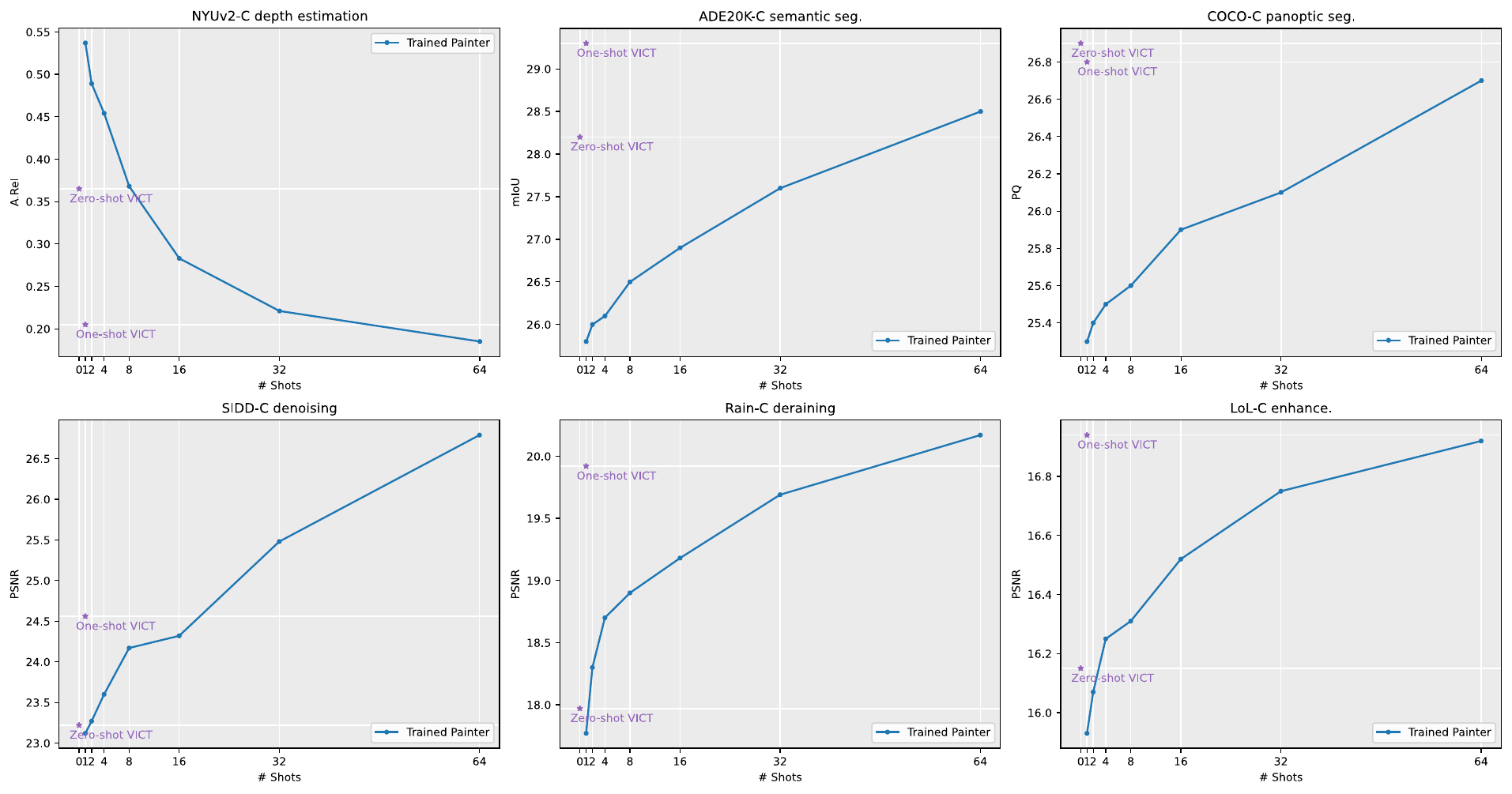}
    \vspace{-5pt}
    \caption{\textbf{Comparison with few-shot Painter on six vision tasks with corruptions.} We randomly corrupt a certain number of images in the training set, using 1, 2, 4, 8, 16, 32, and 64 shots for training and deploying the model in the full corrupted test sets. We report the final results averaged across 15 corruptions. Our zero-shot or one-shot VICT can outperform Painter trained with more few-shot examples.}
    \label{fig:xshot}
    \vspace{-8pt}
\end{figure*}

\subsection{Implementation Details}
\label{subsec:details}

\noindent\textbf{Tasks and datasets.}
Following~\cite{wang2023images}, we evaluate the model performance on six representative vision tasks ranging from high-level visual understanding to low-level image processing, which includes depth estimation on NYUv2~\cite{silberman2012indoor}, semantic segmentation on ADE20K~\cite{zhou2017scene}, panoptic segmentation on COCO~\cite{lin2014microsoft}, image denoising on SIDD~\cite{abdelhamed2018high}, image draining on the merged draining datasets~\cite{zamir2020learning}, and low-light image enhancement on LoL~\cite{wei2018deep}. 
To evaluate the robustness to distribution shifts of the model on these datasets, we follow the setup in~\cite{hendrycks2019benchmarking,michaelis2019benchmarking} to corrupt the aforementioned datasets and simulate hard distribution shifts. The corrupted datasets contain 15 types of corruptions, covering noise (gaussian noise, impulse noise, shot noise), blur (defocus blur, glass blur, motion blur, zoom blur), weather (fog, frost, snow), and digital (brightness, contrast, elastic transform, jpeg compression, pixelate) categories. Each type of corruption has five levels of severity. We denote the corrupted datasets as NYUv2-C, ADE20K-C, COCO-C, SIDD-C, Rain-C, and LoL-C, respectively. Due to the space constraints, the results in the main text are limited to the most severe level (\ie, level 5) that corresponds to the strongest distribution shift. We provide the results on the other four levels in the supplementary material.

\noindent\textbf{Baselines.}
Without loss of generality, we consider Painter~\cite{wang2023images} as our VICL model for its simplicity in design and its wide applicability.
We study VICT in two settings: (\romannumeral1) zero-shot setting, where the task prompts are from the training distribution (\ie, clean images), and (\romannumeral2) one-shot setting, where the task prompts are from the test distribution (\ie, corrupted images).
For the one-shot setting, apart from comparing with the frozen Painter using one-shot corrupted examples as task prompts, we also consider a baseline that further trains Painter with the one-shot corrupted samples using the same pre-training objective as in~\cite{wang2023images}. We also consider training Painter with more few-shot examples using the same settings to examine to what extent VICT can outperform a larger number of few-shot fine-tuning of Painter.

\noindent\textbf{Training details.}
In all experiments, we use the pre-trained Painter based on ViT-Large provided by the authors of~\cite{wang2023images}. We perform VICT using an AdamW~\cite{loshchilov2019decoupled} optimizer, with betas as (0.9, 0.999), a weight decay of 0, a batch size of 1, and a fixed learning rate of 1e-6. Unless otherwise specified, we train each test sample for 60 steps. 
The choice of 60 steps is purely computational and more steps are likely to further improve performance, judging from the positive trend observed in Figure~\ref{fig:steps}.
Note that we do not apply any data augmentations for VICT as many data augmentations are in fact distribution shifts in our evaluation benchmarks. Training with them is analogous to training on the test distributions. Therefore, to solely study the generalization ability to new test distributions, we purposely choose not to use any data augmentations, even though they could improve our results at face value. Every iteration of VICT is performed on the same concatenated grid-like image, with each sub-image resized as $448\times 448$, which is the same as what we later use for in-context inference. During VICT, we only optimize the encoder weights. We have experimented with training both the encoder and the decoder and found that the difference is negligible, which is also consistent with the observations in~\cite{sun2020test,gandelsman2022test}. A detailed comparison is provided in the experiment section.

\subsection{Main Results}
\label{subsec:results}

\begin{table*}[t]
\centering
\small
\caption{\textbf{Test-time optimization on different modules.} We use semantic segmentation on ADE20K-C for the ablation. Training only the encoder performs similarly to training both the encoder and decoder, regardless of the zero-shot and one-shot settings.}
\label{tab:module}
\addtolength{\tabcolsep}{-2pt}
\begin{tabular}{lcccccccccccccccc}
\toprule
module & brigh & cont & defoc & elast & fog & frost & gauss & glass & impul & jpeg & motn & pixel & shot & snow & zoom & avg \\ \midrule
\multicolumn{17}{l}{\emph{zero-shot setting:}} \\
Encoder & 40.9 & 12.7 & 28.2 & 31.7 & 36.4 & 23.7 & 25.5 & 26.3 & 27.4 & 38.8 & 31.9 & 40.9 & 28.0 & 17.3 & 13.6 & 28.2 \\
Both & 40.9 & 12.5 & 28.2 & 31.7 & 36.4 & 23.8 & 25.7 & 26.5 & 27.4 & 38.8 & 31.9 & 41.0 & 27.9 & 16.9 & 13.5 & 28.2 \\ \midrule
\multicolumn{17}{l}{\emph{one-shot setting:}} \\
Encoder & 41.7 & 21.0 & 28.8 & 32.2 & 37.0 & 24.7 & 24.4 & 27.0 & 25.2 & 39.7 & 32.8 & 41.5 & 26.6 & 21.7 & 14.9 & 29.3 \\
Both & 41.7 & 21.0 & 28.7 & 32.1 & 36.9 & 24.4 & 24.4 & 27.0 & 25.3 & 39.6 & 33.0 & 41.6 & 26.8 & 21.5 & 15.1 & 29.3 \\ \bottomrule
\end{tabular}
\end{table*}

\noindent\textbf{System-level comparison.}
We compare VICT with Painter under 15 common corruptions as introduced in Section~\ref{subsec:details}. We perform experiments on six representative vision tasks covering high-level visual understanding and low-level image processing, which includes depth estimation on NYUv2-C, semantic segmentation on ADE20K-C, panoptic segmentation on COCO-C, image denoising on SIDD-C, image draining on Rain-C, and low-light image enhancement on LoL-C.

Table~\ref{tab:results} reports the results. First of all, we observe that Painter exhibits poor generalization ability under common corruptions. For example, on ADE20K-C semantic segmentation, Painter only achieves an average performance of 27.0 mIoU, which is 22.9 mIoU lower than that on the clean validation set (49.9 mIoU as reported in Figure~\ref{fig:teaser}). Nevertheless, our VICT outperforms Painter by clear margins in both zero-shot and one-shot settings across different tasks. When it comes to the zero-shot setting, VICT achieves an average of -0.027 A.Rel on NYUv2-C depth estimation, +1.2 mIoU on ADE20K-C semantic segmentation, +0.5 PQ on COCO-C panoptic segmentation, +0.14 PSNR on SIDD-C image denosing, +0.18 PSNR on Rain-C image draining, and +0.25 PSNR on LoL-C low-light enhancement, respectively. 

The performance gains of VICT over Painter are further increased when we shift from the zero-shot setting to the one-shot counterpart. Specifically, VICT achieves an average of -0.332 A.Rel on NYUv2-C depth estimation, +3.5 mIoU on ADE20K-C semantic segmentation, +1.5 PQ on COCO-C panoptic segmentation, +1.44 PSNR on SIDD-C image denoising, +2.15 PSNR on Rain-C image draining, and +1.01 PSNR on LoL-C low-light enhancement, respectively. Besides, for the Painter baseline, using the one-shot setting does not always perform better than the zero-shot setting. The average performances are even degraded in most cases, \eg, depth estimation on NYUv2-C (0.392 A.Rel \vs 0.537 A.Rel), semantic segmentation on ADE20K-C (27.0 mIoU \vs 25.8 mIoU), panoptic segmentation on COCO-C (26.4 PQ \vs 25.3 PQ), image draining on Rain-C (17.79 PSNR \vs 17.77 PSNR). In contrast, for VICT, using the one-shot setting usually performs better than the zero-shot setting. It is worth mentioning that our zero-shot VICT can even outperform the one-shot Painter, further demonstrating the effectiveness of VICT.

\noindent\textbf{Comparison with few-shot Painter.}
In previous experiments, we mainly compare with a frozen Painter using either clean or corrupted examples as the task prompts. Here, we further consider a scenario where more few-shot examples from the test distribution are available. Specifically, we randomly corrupt a certain number of images in the training set, using 1, 2, 4, 8, 16, 32, and 64 shots to train Painter respectively and deploying the model in the full corrupted test sets.

The results are shown in Figure~\ref{fig:xshot}. Both our zero-shot VICT and one-shot VICT can consistently outperform the one-shot Painter. When more few-shot examples are available for training, our VICT can still outperform Painter below a certain threshold. We observe such a threshold varies for different tasks, with high-level visual understanding tasks having a higher threshold. It is worth noting that in high-level visual understanding tasks like ADE20K-C semantic segmentation and COCO-C panoptic segmentation, both our zero-shot and one-shot variants of VICT nearly match or even outperform the 64-shot Painter. In other words, with only one or simply no labeled test sample(s), VICT can achieve similar performances as models that use significantly more labeled test samples to train. This phenomenon is rather appealing as in practice it is rare to know test distribution in advance, let alone obtaining a large number of labeled test samples.

\subsection{Further Analysis}
\label{subsec:analysis}

\begin{figure}[t]
    \centering
    \includegraphics[width=.99\linewidth]{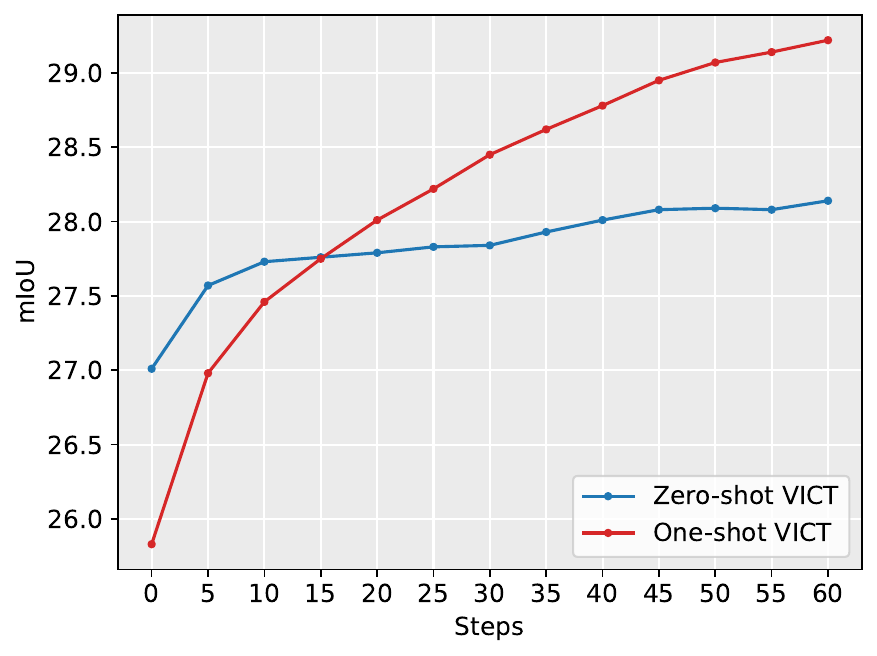}
    \vspace{-5pt}
    \caption{\textbf{Analysis on the trade-off between efficiency and accuracy.} We use semantic segmentation on ADE20K-C for the ablation. VICT benefits from more training steps, while at the cost of linearly increased training time.}
    \label{fig:steps}
    \vspace{-5pt}
\end{figure}

\noindent\textbf{Effect of test-time optimization modules.}
We study the effect of optimizing different modules at test time. We consider two parameter groups: (\romannumeral1) the encoder, and (\romannumeral2) both the encoder and decoder. We use semantic segmentation on ADE20K-C as an example task for analysis. The results are shown in Table~\ref{tab:module}. Training only the encoder performs similarly to training the entire model, regardless of the zero-shot and one-shot settings.
This makes sense since the encoder acts as a feature extractor, which plays a key role in determining the representation quality. In contrast, the decoder is only responsible for mapping the latent representation back to its original resolution, which is more specialized for reconstruction but less relevant for semantics.
Thus, we choose only to optimize the encoder at test time.

\noindent\textbf{Effect of test-time optimization steps.}
We study the effect of test-time optimization steps in Figure~\ref{fig:steps}.
VICT benefits from more training steps, which keeps improving performance even after 60 steps.
However, it should be noted that the runtime of our VICT is linearly increased with the number of training steps. 
For reference, it takes around 0.4 seconds per step per test sample on a single A100 GPU.
Thus, in practice, one may decide the number of training steps according to the pre-defined cost budget for deployment.

\begin{figure}[t]
    \centering
    \includegraphics[width=.99\linewidth]{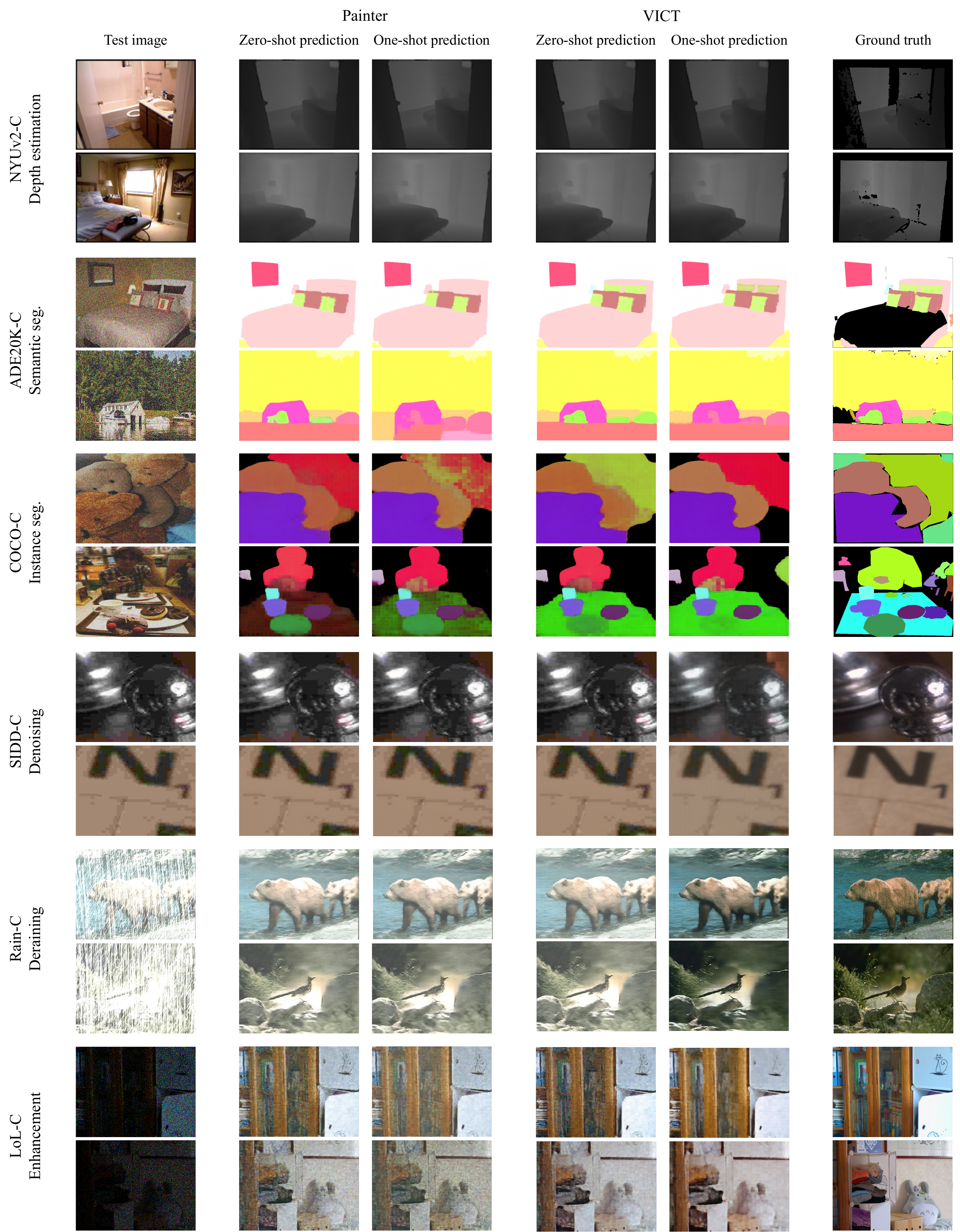}
    \caption{\textbf{Visualizations of test examples and predictions for six main tasks with corruptions.} We visualize both zero-shot and one-shot settings for Painter and VICT. Zoom in for best view.}
    \label{fig:vlz}
    \vspace{-5pt}
\end{figure}

\subsection{Qualitative Results}
\label{subsec:vlz}

\noindent\textbf{Results on main tasks.}
We visualize some test examples and predictions from the validation set for different tasks with corruptions, including depth estimation, semantic segmentation, instance segmentation, image denoising, image deraining, and low-light image enhancement. As shown in Figure~\ref{fig:vlz}, VICT can make more accurate predictions than Painter on all tasks.

\begin{figure}[t]
    \centering
    \includegraphics[width=.99\linewidth]{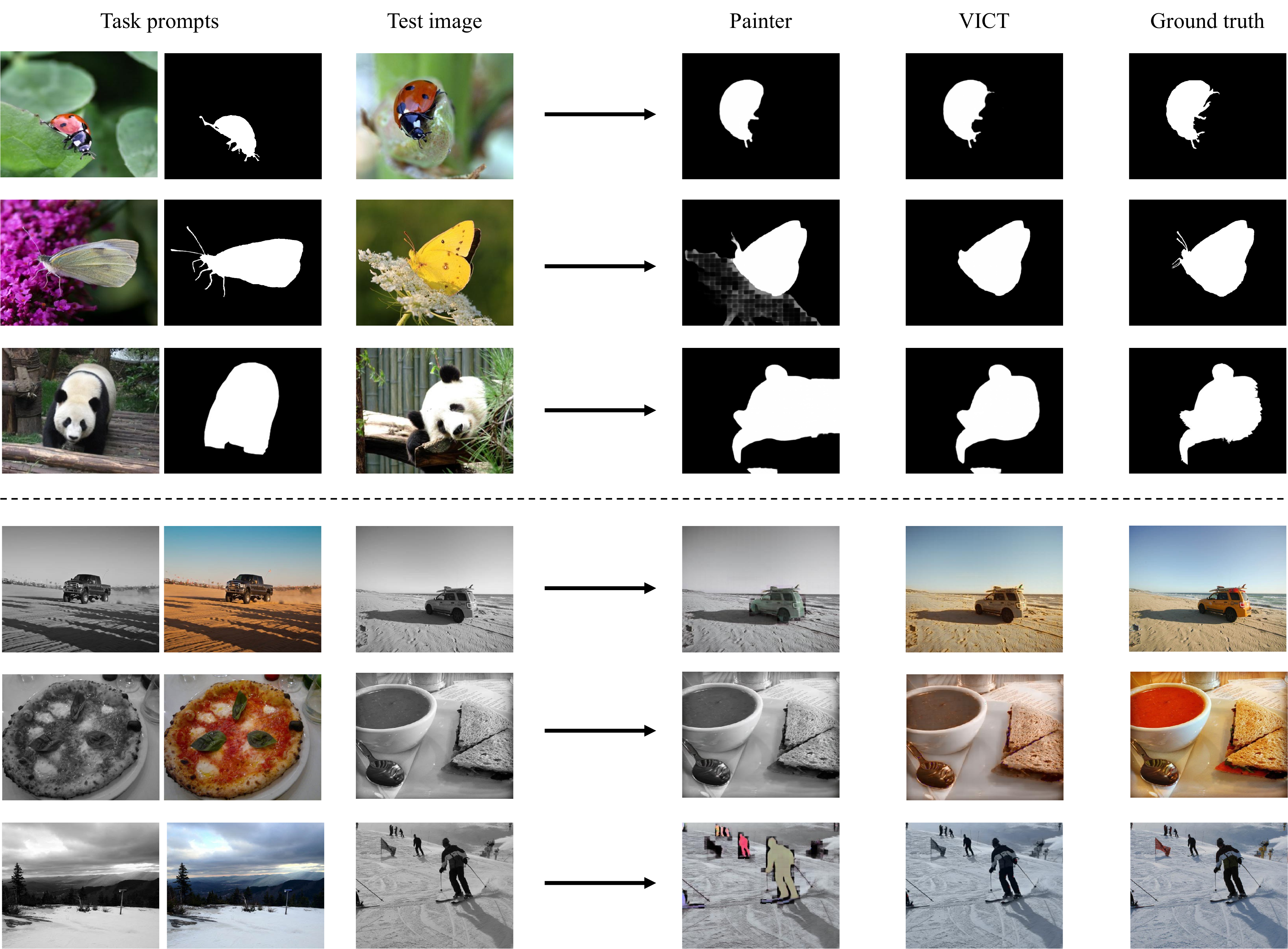}
    \caption{\textbf{Visualizations of test examples and predictions for unseen tasks.} The visualized tasks include foreground object segmentation and colorization. Painter cannot generalize to totally unseen tasks like colorization, whereas VICT can make decent color predictions. Zoom in for best view.}
    \label{fig:vlz_novel}
    \vspace{-5pt}
\end{figure}

\noindent\textbf{Results on unseen tasks.}
In this work, we mainly consider the distribution shifts with corruptions and conduct experiments on the tasks that are seen in training.
Here, we further verify whether VICT can generalize to unseen tasks.
We explore this capability via visualizations.
More quantitative results are provided in the supplementary material.
Figure~\ref{fig:vlz_novel} provides examples of two unseen tasks including foreground object segmentation and colorization. Both Painter and our VICT can generalize to the foreground segmentation task. This makes sense as similar segmentation tasks (\eg, semantic segmentation and instance segmentation) have been seen during the training stage of Painter. Nevertheless, our VICT further reduces noises and produces finer masks. However, Painter cannot generalize to the colorization task that is totally unseen during training, producing only grayscale images or even wrong predictions from other tasks (\eg, the last row). In contrast, our VICT can produce decent colorful results. This further demonstrates the potential of applying VICT for unseen tasks at test time.


\section{Conclusion}
\label{sec:conclusion}

Generalization under distribution shifts is the central theme of deep learning. In this work, we have studied the robustness of VICL models and found that the existing VICL paradigm exhibits poor generalization capability to unseen new domains. Based on this observation, we proposed a simple yet effective test-time visual in-context tuning method to adapt VICL models to a single test sample on the fly. We hope our explorations can pave the way for improving the generalizability of VICL.

\noindent\textbf{Limitations and future work.}
Our study has several limitations: 1) Since we perform training for each test sample, our method is slower than the baseline applying a fixed model at test time, which is a common limitation for test-time training. Inference speed might be improved through better architectural designs, training techniques, optimizers, and hyper-parameters.
However, it has not been the focus of this paper. 
2) Our proposed cycle consistency supervision is a general self-supervised task. However, we cannot guarantee that it will produce useful gradients for every single test distribution. We only focus on several representation vision tasks and most popular corruption benchmarks for distribution shifts. More tasks and benchmarks can be studied for future research.
One possible extension of this work is to apply similar test-time supervision in multi-modal in-context learning, where cycle consistency supervision also exists. We leave this exploration for future work.

\clearpage
{
    \small
    \bibliographystyle{ieeenat_fullname}
    \bibliography{main}
}


\appendix

\twocolumn[{%
\renewcommand\twocolumn[1][]{#1}%
\vspace{+4em}
\maketitlesupplementary
\vspace{+1em}
\begin{center}
\centering
\small
\captionof{table}{\textbf{Effect on in-domain test samples.} We perform VICT on clean test images for each task. VICT does not hurt or only slightly hurts the original performance.}
\label{tab:clean}
\addtolength{\tabcolsep}{-2pt}
\begin{tabular}{lccccccccccc}
\toprule
 & \multicolumn{3}{c}{\begin{tabular}[c]{@{}c@{}}depth estimation\\ NYUv2\end{tabular}} & \begin{tabular}[c]{@{}c@{}}semantic seg.\\ ADE-20K\end{tabular} & \begin{tabular}[c]{@{}c@{}}panoptic seg.\\ COCO\end{tabular} & \multicolumn{2}{c}{\begin{tabular}[c]{@{}c@{}}denoising\\ SIDD\end{tabular}} & \multicolumn{2}{c}{\begin{tabular}[c]{@{}c@{}}deraining\\ 5 datasets\end{tabular}} & \multicolumn{2}{c}{\begin{tabular}[c]{@{}c@{}}enhance.\\ LoL\end{tabular}} \\
 & RMSE $\downarrow$ & A.Rel $\downarrow$ & d1 $\uparrow$ & mIoU $\uparrow$ & PQ $\uparrow$ & PSNR $\uparrow$ & SSIM $\uparrow$ & PSNR $\uparrow$ & SSIM $\uparrow$ & PSNR $\uparrow$ & SSIM $\uparrow$ \\ \midrule
Painter & 0.288 & 0.080 & 0.950 & 49.9 & 43.4 & 38.88 & 0.954 & 29.49 & 0.868 & 22.40 & 0.872 \\
VICT & 0.292 & 0.080 & 0.949 & 49.9 & 43.6 & 38.38 & 0.954 & 29.34 & 0.867 & 22.25 & 0.872 \\ \bottomrule
\end{tabular}
\vspace{+10pt}
\centering
\small
\captionof{table}{\textbf{Results on more VICL baselines and unseen tasks.} VICT can generalize well to more VICL baselines and out-of-domain vision tasks at test time.}
\label{tab:novel}
\addtolength{\tabcolsep}{-2pt}
\begin{tabular}{lcccccccccccccc}
\toprule
 & \multicolumn{5}{c}{\begin{tabular}[c]{@{}c@{}}surface normal estimation\\ Taskonomy\end{tabular}} & \begin{tabular}[c]{@{}c@{}}foreground seg.\\ DUTS\end{tabular} & \multicolumn{2}{c}{\begin{tabular}[c]{@{}c@{}}deblurring\\ GoPro\end{tabular}} & \multicolumn{2}{c}{\begin{tabular}[c]{@{}c@{}}colorization\\ ImageNet\end{tabular}} \\
 & mean $\downarrow$ & median $\downarrow$ & 11.25\textdegree $\uparrow$ & 22.5\textdegree $\uparrow$ & 30\textdegree $\uparrow$ & mIoU $\uparrow$ & PSNR $\uparrow$ & SSIM $\uparrow$ & PSNR $\uparrow$ & SSIM $\uparrow$ \\ \midrule
MAE-VQGAN & 28.8 & 26.8 & 18.6 & 41.4 & 54.5 & 38.1 & 14.05 & 0.559 & 13.60 & 0.481 \\
w/ VICT & 25.1 & 23.1 & 21.0 & 48.0 & 66.3 & 41.1 & 14.27 & 0.569 & 14.17 & 0.509 \\ \midrule
Painter & 37.2 & 36.1 & 2.6 & 20.0 & 44.2 & 48.5 & 23.88 & 0.818 & 18.14 & 0.809 \\
w/ VICT & 23.1 & 21.0 & 26.4 & 52.9 & 69.9 & 54.1 & 24.50 & 0.829 & 20.04 & 0.873 \\ \bottomrule
\end{tabular}
\vspace{+10pt}
\end{center}%
}]

\section{Effect on In-Domain Test Samples}

In the main text, we apply VICT when a distribution shift occurs at test time. It is interesting to ask whether VICT works when the training and test distributions are the same.
To study this, we perform VICT on clean test images for each task. As shown in Table~\ref{tab:clean}, we observe no further performance gains or slight performance degradations compared with the Painter baseline.
Since the Painter model has been well pre-trained on clean images, further applying VICT cannot extract additional information from data belonging to the training distribution and might instead lead to slight overfitting in some cases.
In practice, one may rely on prior domain knowledge about the deployed environment to determine whether to apply VICT. Even if such knowledge is unavailable, in most cases, VICT does not hurt or only slightly hurts the original performance, but can significantly improve on many new distributions.

\section{Results on More Baselines and Tasks}

In the main text, we mainly consider Painter as our baseline. To further demonstrate the generalizability of our method, we apply VICT to another VICL baseline, \ie, MAE-VQGAN~\cite{bar2022visual}. Besides, we also apply VICT to more unseen tasks, including surface normal estimation on Taskonomy~\cite{zamir2018taskonomy}, foreground segmentation on DUTS~\cite{wang2017learning}, image deblurring on GoPro~\cite{nah2017deep}, and colorization on ImageNet~\cite{deng2009imagenet}. We randomly sample 1000 images from their respective test or validation sets for evaluation. We sweep the learning rate for different tasks and train each test sample for 100 steps. As shown in Table~\ref{tab:novel}, VICT consistently outperforms each baseline on different tasks by clear margins.

\section{Results on Other Severity Levels}

Table~\ref{tab:results_level1}-\ref{tab:results_level4} present the results on the other four corruption levels.
In all experiments, we perform VICT using the same hyper-parameters as described in Section~\ref{subsec:details} in the main text without specialized tuning, except that we train each test sample for 20 steps (instead of 60, for faster experiments). Note that using the same hyper-parameters across different corruptions and levels can be sub-optimal. Nevertheless, VICT outperforms Painter regardless of settings, tasks, and corruption levels.

\clearpage
\begin{table*}[t]
\centering
\small
\caption{\textbf{System-level comparison on six representative vision tasks across 15 corruptions.} Results are on corruption level 1.}
\label{tab:results_level1}
\addtolength{\tabcolsep}{-2pt}
\begin{tabular}{lcccccccccccccccc}
\toprule
method & brigh & cont & defoc & elast & fog & frost & gauss & glass & impul & jpeg & motn & pixel & shot & snow & zoom & avg \\ \hline\midrule
\multicolumn{17}{c}{(a) depth estimation NYUv2-C (A.Rel $\downarrow$)} \\ \midrule
\multicolumn{17}{l}{\emph{zero-shot setting:}} \\
Painter & 0.094 & 0.089 & 0.349 & 0.096 & 0.093 & 0.120 & 0.096 & 0.250 & 0.109 & 0.169 & 0.103 & 0.100 & 0.093 & 0.209 & 0.154 & 0.142 \\
VICT & 0.083 & 0.101 & 0.295 & 0.085 & 0.105 & 0.120 & 0.089 & 0.202 & 0.095 & 0.133 & 0.089 & 0.085 & 0.085 & 0.209 & 0.127 & 0.127 \\ \midrule
\multicolumn{17}{l}{\emph{one-shot setting:}} \\
Painter & 0.093 & 0.089 & 0.351 & 0.097 & 0.093 & 0.136 & 0.099 & 0.248 & 0.113 & 0.170 & 0.101 & 0.101 & 0.097 & 0.270 & 0.152 & 0.147 \\
VICT & 0.084 & 0.093 & 0.098 & 0.084 & 0.096 & 0.114 & 0.090 & 0.088 & 0.094 & 0.089 & 0.087 & 0.084 & 0.086 & 0.130 & 0.117 & 0.096 \\ \hline\midrule
\multicolumn{17}{c}{(b) semantic segmentation ADE20K-C (mIoU $\uparrow$)} \\ \midrule
\multicolumn{17}{l}{\emph{zero-shot setting:}} \\
Painter & 48.9 & 46.5 & 45.3 & 45.8 & 45.9 & 43.1 & 46.4 & 45.4 & 46.1 & 48.4 & 47.8 & 48.9 & 46.2 & 43.8 & 31.4 & 45.3 \\
VICT & 49.1 & 46.6 & 45.6 & 45.9 & 46.4 & 43.4 & 46.9 & 45.5 & 46.7 & 48.3 & 47.6 & 48.9 & 47.5 & 43.8 & 31.9 & 45.6 \\ \midrule
\multicolumn{17}{l}{\emph{one-shot setting:}} \\
Painter & 48.9 & 46.4 & 45.1 & 45.9 & 45.9 & 43.2 & 46.6 & 45.6 & 45.9 & 48.4 & 47.7 & 48.9 & 46.3 & 42.8 & 31.5 & 45.3 \\
VICT & 49.2 & 46.7 & 45.3 & 46.1 & 46.0 & 44.4 & 47.6 & 46.0 & 47.3 & 48.5 & 48.0 & 48.7 & 47.5 & 43.3 & 31.9 & 45.8 \\ \hline\midrule
\multicolumn{17}{c}{(c) panoptic segmentation COCO-C (PQ $\uparrow$)} \\ \midrule
\multicolumn{17}{l}{\emph{zero-shot setting:}} \\
Painter & 42.2 & 40.7 & 39.5 & 40.3 & 40.0 & 38.1 & 40.2 & 39.5 & 39.2 & 40.2 & 41.0 & 41.2 & 40.4 & 37.6 & 27.7 & 39.2 \\
VICT & 42.9 & 41.2 & 40.0 & 40.7 & 40.5 & 38.6 & 40.8 & 40.3 & 39.8 & 40.8 & 41.3 & 42.0 & 40.9 & 38.2 & 28.2 & 39.7 \\ \midrule
\multicolumn{17}{l}{\emph{one-shot setting:}} \\
Painter & 42.1 & 40.3 & 39.3 & 40.1 & 39.5 & 38.0 & 40.0 & 39.4 & 38.8 & 39.4 & 40.9 & 41.0 & 40.2 & 37.2 & 27.5 & 38.9 \\
VICT & 42.8 & 40.9 & 39.7 & 40.6 & 40.3 & 38.6 & 40.6 & 40.0 & 39.4 & 40.5 & 41.2 & 41.8 & 40.9 & 37.6 & 27.8 & 39.5 \\ \hline\midrule
\multicolumn{17}{c}{(d) denoising SIDD-C (PSNR $\uparrow$)} \\ \midrule
\multicolumn{17}{l}{\emph{zero-shot setting:}} \\
Painter & 22.53 & 28.35 & 34.64 & 36.06 & 18.72 & 12.34 & 35.63 & 33.61 & 34.35 & 32.88 & 33.88 & 37.36 & 37.01 & 16.19 & 31.37 & 29.66 \\
VICT & 22.90 & 28.50 & 34.49 & 35.87 & 18.85 & 12.50 & 35.59 & 33.48 & 34.42 & 32.86 & 33.75 & 37.24 & 37.06 & 16.27 & 31.36 & 29.68 \\ \midrule
\multicolumn{17}{l}{\emph{one-shot setting:}} \\
Painter & 23.20 & 28.36 & 34.95 & 36.09 & 18.56 & 12.98 & 35.87 & 33.89 & 34.61 & 32.60 & 33.98 & 37.36 & 37.09 & 16.88 & 31.38 & 29.85 \\
VICT & 28.07 & 27.82 & 35.04 & 36.27 & 17.68 & 15.77 & 35.65 & 34.46 & 35.21 & 33.09 & 34.43 & 37.00 & 36.45 & 21.14 & 31.92 & 30.67 \\ \hline\midrule
\multicolumn{17}{c}{(e) deraining Rain-C (PSNR $\uparrow$)} \\ \midrule
\multicolumn{17}{l}{\emph{zero-shot setting:}} \\
Painter & 22.94 & 18.04 & 24.17 & 24.39 & 15.95 & 14.84 & 27.50 & 23.91 & 26.94 & 28.16 & 23.61 & 27.62 & 27.22 & 17.95 & 19.62 & 22.86 \\
VICT & 23.03 & 18.10 & 24.23 & 24.39 & 16.00 & 14.96 & 27.63 & 23.93 & 27.02 & 28.12 & 23.62 & 27.58 & 27.37 & 18.11 & 19.67 & 22.92 \\ \midrule
\multicolumn{17}{l}{\emph{one-shot setting:}} \\
Painter & 23.34 & 18.20 & 24.10 & 24.39 & 15.97 & 15.24 & 27.43 & 23.94 & 26.87 & 28.16 & 23.65 & 27.63 & 27.21 & 18.30 & 19.65 & 22.94 \\
VICT & 27.94 & 20.48 & 24.53 & 24.49 & 16.18 & 21.35 & 27.63 & 24.13 & 27.15 & 28.14 & 23.92 & 27.59 & 27.51 & 23.86 & 20.02 & 24.33 \\ \hline\midrule
\multicolumn{17}{c}{(f) low-light enhancement LoL-C (PSNR $\uparrow$)} \\ \midrule
\multicolumn{17}{l}{\emph{zero-shot setting:}} \\
Painter & 18.91 & 17.58 & 21.50 & 21.25 & 12.43 & 14.20 & 17.00 & 20.79 & 18.57 & 19.07 & 20.78 & 21.92 & 21.51 & 12.45 & 18.42 & 18.43 \\
VICT & 19.00 & 17.75 & 21.53 & 21.23 & 12.53 & 14.26 & 17.48 & 20.76 & 19.25 & 19.11 & 20.83 & 21.94 & 21.47 & 12.58 & 18.43 & 18.54 \\ \midrule
\multicolumn{17}{l}{\emph{one-shot setting:}} \\
\multicolumn{1}{l}{Painter} & 19.27 & 17.84 & 21.50 & 21.24 & 12.33 & 14.20 & 16.69 & 20.81 & 16.60 & 19.08 & 20.78 & 21.92 & 21.55 & 11.62 & 18.41 & 18.26 \\
VICT & 20.72 & 19.65 & 21.26 & 21.16 & 12.53 & 14.93 & 19.74 & 20.79 & 19.41 & 19.30 & 20.74 & 21.90 & 21.56 & 15.45 & 18.37 & 19.17 \\ \bottomrule
\end{tabular}
\end{table*}

\begin{table*}[t]
\centering
\small
\caption{\textbf{System-level comparison on six representative vision tasks across 15 corruptions.} Results are on corruption level 2.}
\label{tab:results_level2}
\addtolength{\tabcolsep}{-2pt}
\begin{tabular}{lcccccccccccccccc}
\toprule
method & brigh & cont & defoc & elast & fog & frost & gauss & glass & impul & jpeg & motn & pixel & shot & snow & zoom & avg \\ \hline\midrule
\multicolumn{17}{c}{(a) depth estimation NYUv2-C (A.Rel $\downarrow$)} \\ \midrule
\multicolumn{17}{l}{\emph{zero-shot setting:}} \\
Painter & 0.101 & 0.097 & 0.447 & 0.097 & 0.100 & 0.255 & 0.104 & 0.333 & 0.107 & 0.214 & 0.114 & 0.093 & 0.097 & 0.939 & 0.168 & 0.218 \\
VICT & 0.088 & 0.114 & 0.390 & 0.087 & 0.113 & 0.255 & 0.096 & 0.281 & 0.099 & 0.173 & 0.096 & 0.084 & 0.091 & 0.939 & 0.143 & 0.203 \\ \midrule
\multicolumn{17}{l}{\emph{one-shot setting:}} \\
Painter & 0.100 & 0.097 & 0.457 & 0.098 & 0.100 & 0.376 & 0.117 & 0.330 & 0.131 & 0.212 & 0.112 & 0.094 & 0.104 & 1.412 & 0.168 & 0.261 \\
VICT & 0.086 & 0.099 & 0.106 & 0.088 & 0.102 & 0.155 & 0.099 & 0.095 & 0.106 & 0.096 & 0.092 & 0.083 & 0.095 & 0.625 & 0.136 & 0.138 \\ \hline\midrule
\multicolumn{17}{c}{(b) semantic segmentation ADE20K-C (mIoU $\uparrow$)} \\ \midrule
\multicolumn{17}{l}{\emph{zero-shot setting:}} \\
Painter & 47.1 & 45.0 & 42.6 & 43.3 & 44.2 & 35.7 & 44.1 & 43.3 & 43.5 & 47.5 & 44.9 & 48.3 & 43.7 & 29.2 & 25.8 & 41.9 \\
VICT & 47.5 & 45.5 & 42.9 & 42.8 & 45.2 & 35.9 & 44.2 & 43.4 & 43.6 & 47.4 & 45.6 & 48.6 & 43.9 & 30.8 & 26.1 & 42.2 \\ \midrule
\multicolumn{17}{l}{\emph{one-shot setting:}} \\
Painter & 47.1 & 45.0 & 42.4 & 43.3 & 44.2 & 35.3 & 43.8 & 43.4 & 43.0 & 47.5 & 44.7 & 48.2 & 43.9 & 23.2 & 26.0 & 41.4 \\
VICT & 47.8 & 45.4 & 43.4 & 43.1 & 44.6 & 36.6 & 44.7 & 43.6 & 43.8 & 47.6 & 44.9 & 48.6 & 44.3 & 32.7 & 26.7 & 42.5 \\ \hline\midrule
\multicolumn{17}{c}{(c) panoptic segmentation COCO-C (PQ $\uparrow$)} \\ \midrule
\multicolumn{17}{l}{\emph{zero-shot setting:}} \\
Painter & 41.5 & 39.5 & 37.6 & 38.5 & 38.8 & 33.0 & 38.5 & 37.4 & 37.3 & 38.9 & 38.8 & 41.2 & 38.4 & 30.4 & 23.4 & 36.9 \\
VICT & 42.2 & 40.0 & 38.0 & 39.1 & 39.5 & 33.6 & 39.1 & 37.9 & 37.8 & 39.6 & 39.0 & 41.8 & 39.0 & 31.4 & 23.5 & 37.4 \\ \midrule
\multicolumn{17}{l}{\emph{one-shot setting:}} \\
Painter & 41.4 & 38.9 & 37.2 & 38.4 & 38.2 & 32.5 & 38.0 & 37.2 & 36.7 & 38.2 & 38.7 & 41.0 & 38.0 & 29.2 & 23.0 & 36.4 \\
VICT & 42.0 & 39.6 & 37.6 & 39.0 & 39.1 & 33.3 & 38.6 & 37.8 & 37.3 & 39.1 & 38.8 & 41.7 & 38.6 & 31.5 & 23.5 & 37.2 \\ \hline\midrule
\multicolumn{17}{c}{(d) denoising SIDD-C (PSNR $\uparrow$)} \\ \midrule
\multicolumn{17}{l}{\emph{zero-shot setting:}} \\
Painter & 16.32 & 27.50 & 34.50 & 35.34 & 17.96 & 9.84 & 31.31 & 33.58 & 29.81 & 30.56 & 32.61 & 37.48 & 34.69 & 11.97 & 30.61 & 27.61 \\
VICT & 16.54 & 27.64 & 34.40 & 35.09 & 18.13 & 9.90 & 31.34 & 33.40 & 29.74 & 31.93 & 32.47 & 37.36 & 34.73 & 11.93 & 30.57 & 27.68 \\ \midrule
\multicolumn{17}{l}{\emph{one-shot setting:}} \\
Painter & 17.05 & 27.51 & 34.70 & 35.33 & 17.63 & 10.39 & 32.87 & 33.94 & 30.78 & 30.23 & 32.68 & 37.46 & 35.87 & 12.62 & 30.61 & 27.98 \\
VICT & 21.52 & 27.20 & 34.86 & 35.70 & 16.42 & 11.59 & 33.24 & 34.34 & 32.05 & 32.67 & 33.48 & 37.14 & 35.54 & 15.90 & 31.10 & 28.85 \\ \hline\midrule
\multicolumn{17}{c}{(e) deraining Rain-C (PSNR $\uparrow$)} \\ \midrule
\multicolumn{17}{l}{\emph{zero-shot setting:}} \\
Painter & 17.95 & 16.75 & 22.96 & 23.22 & 15.10 & 13.23 & 26.36 & 23.18 & 25.57 & 27.68 & 21.60 & 27.14 & 25.62 & 14.45 & 18.75 & 21.30 \\
VICT & 18.10 & 16.81 & 23.03 & 23.27 & 15.15 & 13.34 & 26.54 & 23.28 & 25.77 & 27.68 & 21.67 & 27.14 & 25.91 & 14.63 & 18.80 & 21.41 \\ \midrule
\multicolumn{17}{l}{\emph{one-shot setting:}} \\
Painter & 18.39 & 16.85 & 22.88 & 23.22 & 15.05 & 13.66 & 26.25 & 23.20 & 25.43 & 27.69 & 21.60 & 27.16 & 25.54 & 14.80 & 18.79 & 21.37 \\
VICT & 24.52 & 18.27 & 23.51 & 23.55 & 15.19 & 19.15 & 26.68 & 23.89 & 26.07 & 27.67 & 21.99 & 27.18 & 26.16 & 21.09 & 19.15 & 22.94 \\ \hline\midrule
\multicolumn{17}{c}{(f) low-light enhancement LoL-C (PSNR $\uparrow$)} \\ \midrule
\multicolumn{17}{l}{\emph{zero-shot setting:}} \\
Painter & 17.51 & 16.58 & 21.05 & 20.83 & 12.35 & 13.26 & 14.23 & 20.80 & 15.96 & 18.91 & 19.91 & 21.78 & 20.37 & 12.77 & 17.72 & 17.60 \\
VICT & 17.66 & 16.72 & 21.16 & 20.77 & 12.64 & 13.24 & 14.68 & 20.89 & 16.58 & 18.88 & 19.95 & 21.82 & 20.39 & 12.88 & 17.75 & 17.73 \\ \midrule
\multicolumn{17}{l}{\emph{one-shot setting:}} \\
\multicolumn{1}{l}{Painter} & 17.89 & 16.86 & 21.07 & 20.83 & 12.04 & 13.30 & 14.07 & 20.83 & 14.57 & 18.92 & 19.93 & 21.79 & 20.39 & 12.09 & 17.71 & 17.49 \\
VICT & 19.29 & 19.16 & 20.91 & 20.81 & 12.55 & 13.20 & 16.16 & 20.75 & 17.03 & 18.59 & 19.94 & 21.77 & 20.74 & 13.89 & 17.74 & 18.17 \\ \bottomrule
\end{tabular}
\end{table*}

\begin{table*}[t]
\centering
\small
\caption{\textbf{System-level comparison on six representative vision tasks across 15 corruptions.} Results are on corruption level 3.}
\label{tab:results_level3}
\addtolength{\tabcolsep}{-2pt}
\begin{tabular}{lcccccccccccccccc}
\toprule
method & brigh & cont & defoc & elast & fog & frost & gauss & glass & impul & jpeg & motn & pixel & shot & snow & zoom & avg \\ \hline\midrule
\multicolumn{17}{c}{(a) depth estimation NYUv2-C (A.Rel $\downarrow$)} \\ \midrule
\multicolumn{17}{l}{\emph{zero-shot setting:}} \\
Painter & 0.109 & 0.112 & 0.550 & 0.099 & 0.106 & 0.423 & 0.114 & 0.390 & 0.112 & 0.248 & 0.127 & 0.134 & 0.102 & 1.257 & 0.164 & 0.270 \\
VICT & 0.093 & 0.133 & 0.491 & 0.092 & 0.122 & 0.423 & 0.111 & 0.332 & 0.109 & 0.204 & 0.105 & 0.105 & 0.100 & 1.257 & 0.155 & 0.255 \\ \midrule
\multicolumn{17}{l}{\emph{one-shot setting:}} \\
Painter & 0.107 & 0.116 & 0.569 & 0.100 & 0.107 & 0.608 & 0.212 & 0.391 & 0.222 & 0.247 & 0.125 & 0.137 & 0.128 & 1.528 & 0.164 & 0.317 \\
VICT & 0.088 & 0.117 & 0.125 & 0.092 & 0.108 & 0.185 & 0.129 & 0.125 & 0.134 & 0.105 & 0.098 & 0.091 & 0.108 & 0.795 & 0.162 & 0.164 \\ \hline\midrule
\multicolumn{17}{c}{(b) semantic segmentation ADE20K-C (mIoU $\uparrow$)} \\ \midrule
\multicolumn{17}{l}{\emph{zero-shot setting:}} \\
Painter & 46.4 & 41.1 & 36.7 & 39.8 & 42.0 & 27.9 & 39.8 & 33.2 & 40.1 & 45.8 & 41.2 & 46.3 & 39.5 & 24.1 & 21.2 & 37.7 \\
VICT & 46.4 & 41.4 & 37.2 & 40.2 & 42.3 & 28.7 & 39.9 & 34.1 & 41.1 & 45.7 & 41.0 & 46.7 & 39.9 & 28.4 & 21.9 & 38.3 \\ \midrule
\multicolumn{17}{l}{\emph{one-shot setting:}} \\
Painter & 46.5 & 41.4 & 36.5 & 39.9 & 42.4 & 26.6 & 38.5 & 33.4 & 39.1 & 45.7 & 41.2 & 46.4 & 38.9 & 18.9 & 21.3 & 37.1 \\
VICT & 46.5 & 42.1 & 37.4 & 40.7 & 42.8 & 28.8 & 39.4 & 34.6 & 40.7 & 46.2 & 41.1 & 47.0 & 40.2 & 33.9 & 22.4 & 38.9 \\ \hline\midrule
\multicolumn{17}{c}{(c) panoptic segmentation COCO-C (PQ $\uparrow$)} \\ \midrule
\multicolumn{17}{l}{\emph{zero-shot setting:}} \\
Painter & 40.7 & 36.9 & 33.0 & 35.9 & 37.5 & 29.3 & 35.7 & 29.7 & 35.6 & 38.1 & 35.2 & 39.5 & 35.8 & 28.3 & 19.0 & 34.0 \\
VICT & 41.3 & 37.4 & 33.5 & 36.3 & 38.0 & 29.8 & 36.2 & 30.3 & 36.0 & 38.7 & 35.4 & 40.0 & 36.4 & 29.8 & 19.3 & 34.6 \\ \midrule
\multicolumn{17}{l}{\emph{one-shot setting:}} \\
Painter & 40.7 & 36.4 & 32.7 & 35.7 & 37.1 & 28.7 & 34.8 & 29.6 & 34.7 & 37.4 & 35.2 & 39.3 & 35.0 & 26.5 & 18.5 & 33.5 \\
VICT & 41.1 & 36.9 & 33.2 & 36.3 & 37.6 & 29.7 & 35.4 & 30.4 & 35.4 & 38.2 & 35.2 & 39.7 & 35.8 & 30.8 & 19.0 & 34.3 \\ \hline\midrule
\multicolumn{17}{c}{(d) denoising SIDD-C (PSNR $\uparrow$)} \\ \midrule
\multicolumn{17}{l}{\emph{zero-shot setting:}} \\
Painter & 12.83 & 26.95 & 34.34 & 34.39 & 17.35 & 8.90 & 25.99 & 32.31 & 26.27 & 29.35 & 31.48 & 36.30 & 31.45 & 11.63 & 30.38 & 25.99 \\
VICT & 13.00 & 26.95 & 34.18 & 34.13 & 17.48 & 8.94 & 26.10 & 32.26 & 26.24 & 30.96 & 31.32 & 36.18 & 31.48 & 11.59 & 30.33 & 26.08 \\ \midrule
\multicolumn{17}{l}{\emph{one-shot setting:}} \\
Painter & 13.47 & 26.97 & 34.37 & 34.42 & 17.06 & 9.36 & 27.79 & 32.57 & 27.93 & 29.07 & 31.48 & 36.28 & 32.36 & 12.14 & 30.34 & 26.37 \\
VICT & 16.80 & 26.86 & 34.10 & 35.18 & 15.86 & 11.07 & 29.68 & 33.13 & 29.28 & 31.83 & 32.10 & 36.10 & 33.31 & 15.85 & 30.67 & 27.45 \\ \hline\midrule
\multicolumn{17}{c}{(e) deraining Rain-C (PSNR $\uparrow$)} \\ \midrule
\multicolumn{17}{l}{\emph{zero-shot setting:}} \\
Painter & 14.69 & 15.62 & 21.22 & 21.91 & 14.33 & 12.27 & 24.59 & 20.35 & 24.50 & 27.31 & 20.04 & 25.35 & 23.77 & 14.93 & 18.57 & 19.96 \\
VICT & 14.81 & 15.69 & 21.28 & 22.02 & 14.39 & 12.40 & 24.90 & 20.47 & 24.80 & 27.29 & 20.11 & 25.39 & 24.20 & 15.17 & 18.61 & 20.10 \\ \midrule
\multicolumn{17}{l}{\emph{one-shot setting:}} \\
Painter & 15.09 & 15.64 & 21.20 & 21.92 & 14.28 & 12.66 & 24.47 & 20.42 & 24.27 & 27.32 & 20.11 & 25.40 & 23.53 & 15.24 & 18.60 & 20.01 \\
VICT & 20.81 & 16.43 & 21.92 & 22.55 & 14.57 & 18.01 & 25.33 & 21.43 & 25.26 & 27.37 & 20.71 & 25.58 & 24.38 & 20.93 & 19.01 & 21.62 \\ \hline\midrule
\multicolumn{17}{c}{(f) low-light enhancement LoL-C (PSNR $\uparrow$)} \\ \midrule
\multicolumn{17}{l}{\emph{zero-shot setting:}} \\
Painter & 17.00 & 15.78 & 20.14 & 20.43 & 10.93 & 12.45 & 12.64 & 19.55 & 14.44 & 18.38 & 19.02 & 21.36 & 20.15 & 12.36 & 17.40 & 16.80 \\
VICT & 17.12 & 15.85 & 20.17 & 20.42 & 10.93 & 12.44 & 12.58 & 19.60 & 14.60 & 18.31 & 19.07 & 21.36 & 20.43 & 12.56 & 17.49 & 16.86 \\ \midrule
\multicolumn{17}{l}{\emph{one-shot setting:}} \\
\multicolumn{1}{l}{Painter} & 17.36 & 16.00 & 20.16 & 20.43 & 10.56 & 12.88 & 12.59 & 19.58 & 13.99 & 18.28 & 19.07 & 21.36 & 19.97 & 11.66 & 17.38 & 16.75 \\
VICT & 18.34 & 17.85 & 20.19 & 20.49 & 11.44 & 13.09 & 13.07 & 19.59 & 16.64 & 17.95 & 19.11 & 21.38 & 20.73 & 13.35 & 17.47 & 17.38 \\ \bottomrule
\end{tabular}
\end{table*}

\begin{table*}[t]
\centering
\small
\caption{\textbf{System-level comparison on six representative vision tasks across 15 corruptions.} Results are on corruption level 4.}
\label{tab:results_level4}
\addtolength{\tabcolsep}{-2pt}
\begin{tabular}{lcccccccccccccccc}
\toprule
method & brigh & cont & defoc & elast & fog & frost & gauss & glass & impul & jpeg & motn & pixel & shot & snow & zoom & avg \\ \hline\midrule
\multicolumn{17}{c}{(a) depth estimation NYUv2-C (A.Rel $\downarrow$)} \\ \midrule
\multicolumn{17}{l}{\emph{zero-shot setting:}} \\
Painter & 0.118 & 0.148 & 0.646 & 0.102 & 0.111 & 0.449 & 0.141 & 0.440 & 0.139 & 0.325 & 0.154 & 0.178 & 0.120 & 1.500 & 0.185 & 0.317 \\
VICT & 0.100 & 0.170 & 0.582 & 0.098 & 0.128 & 0.449 & 0.137 & 0.380 & 0.139 & 0.275 & 0.128 & 0.136 & 0.120 & 1.500 & 0.178 & 0.301 \\ \midrule
\multicolumn{17}{l}{\emph{one-shot setting:}} \\
Painter & 0.115 & 0.185 & 0.668 & 0.103 & 0.112 & 0.696 & 0.470 & 0.443 & 0.568 & 0.324 & 0.157 & 0.180 & 0.263 & 1.699 & 0.186 & 0.411 \\
VICT & 0.092 & 0.164 & 0.181 & 0.098 & 0.114 & 0.216 & 0.173 & 0.174 & 0.247 & 0.127 & 0.114 & 0.103 & 0.147 & 1.017 & 0.189 & 0.210 \\ \hline\midrule
\multicolumn{17}{c}{(b) semantic segmentation ADE20K-C (mIoU $\uparrow$)} \\ \midrule
\multicolumn{17}{l}{\emph{zero-shot setting:}} \\
Painter & 43.4 & 28.4 & 32.1 & 35.6 & 40.9 & 27.8 & 33.7 & 30.5 & 34.0 & 43.1 & 35.2 & 43.5 & 31.9 & 16.3 & 17.5 & 32.9 \\
VICT & 44.0 & 28.9 & 32.2 & 36.0 & 41.0 & 28.6 & 34.0 & 30.9 & 34.7 & 43.0 & 36.0 & 43.6 & 33.0 & 20.6 & 17.8 & 33.6 \\ \midrule
\multicolumn{17}{l}{\emph{one-shot setting:}} \\
Painter & 44.1 & 28.6 & 31.6 & 35.8 & 41.2 & 27.1 & 30.9 & 30.8 & 30.7 & 43.0 & 35.4 & 43.5 & 30.5 & 12.3 & 17.6 & 32.2 \\
VICT & 44.4 & 31.5 & 32.7 & 35.6 & 41.2 & 29.4 & 33.0 & 31.6 & 33.3 & 43.6 & 36.3 & 44.3 & 32.8 & 19.3 & 18.6 & 33.8 \\ \hline\midrule
\multicolumn{17}{c}{(c) panoptic segmentation COCO-C (PQ $\uparrow$)} \\ \midrule
\multicolumn{17}{l}{\emph{zero-shot setting:}} \\
Painter & 39.5 & 28.5 & 29.4 & 33.2 & 36.7 & 28.7 & 31.3 & 27.7 & 31.2 & 35.8 & 30.6 & 37.0 & 30.7 & 23.7 & 15.6 & 30.6 \\
VICT & 40.2 & 29.2 & 29.6 & 33.6 & 37.2 & 29.2 & 31.8 & 28.3 & 31.5 & 36.3 & 30.7 & 37.5 & 31.0 & 25.1 & 15.9 & 31.1 \\ \midrule
\multicolumn{17}{l}{\emph{one-shot setting:}} \\
Painter & 39.6 & 28.0 & 28.8 & 33.1 & 36.2 & 27.9 & 29.7 & 27.4 & 29.3 & 35.0 & 30.5 & 36.6 & 28.8 & 21.8 & 15.3 & 29.9 \\
VICT & 40.0 & 28.7 & 29.3 & 33.8 & 36.8 & 29.2 & 30.8 & 28.1 & 30.5 & 36.0 & 30.5 & 37.1 & 29.9 & 26.6 & 15.7 & 30.9 \\ \hline\midrule
\multicolumn{17}{c}{(d) denoising SIDD-C (PSNR $\uparrow$)} \\ \midrule
\multicolumn{17}{l}{\emph{zero-shot setting:}} \\
Painter & 10.41 & 26.06 & 33.67 & 33.80 & 17.21 & 9.15 & 21.44 & 32.94 & 20.98 & 25.92 & 30.22 & 35.54 & 27.26 & 9.54 & 30.01 & 24.28 \\
VICT & 10.55 & 26.01 & 33.69 & 33.54 & 17.31 & 9.20 & 21.53 & 32.86 & 20.95 & 27.95 & 30.31 & 35.43 & 27.27 & 9.49 & 29.94 & 24.40 \\ \midrule
\multicolumn{17}{l}{\emph{one-shot setting:}} \\
Painter & 10.91 & 25.83 & 33.38 & 33.83 & 16.94 & 9.54 & 22.65 & 33.17 & 22.37 & 25.64 & 30.12 & 35.56 & 27.02 & 9.92 & 29.97 & 24.46 \\
VICT & 13.95 & 26.06 & 33.02 & 34.87 & 15.93 & 12.07 & 24.85 & 33.31 & 23.88 & 28.74 & 30.86 & 35.52 & 27.43 & 11.89 & 30.25 & 25.51 \\ \hline\midrule
\multicolumn{17}{c}{(e) deraining Rain-C (PSNR $\uparrow$)} \\ \midrule
\multicolumn{17}{l}{\emph{zero-shot setting:}} \\
Painter & 12.37 & 14.63 & 20.13 & 21.07 & 14.30 & 12.98 & 22.35 & 20.30 & 22.27 & 26.23 & 19.00 & 23.91 & 21.32 & 13.54 & 18.07 & 18.83 \\
VICT & 12.49 & 14.70 & 20.16 & 21.24 & 14.35 & 13.11 & 22.85 & 20.38 & 22.75 & 26.27 & 19.05 & 24.06 & 21.72 & 13.82 & 18.11 & 19.00 \\ \midrule
\multicolumn{17}{l}{\emph{one-shot setting:}} \\
Painter & 12.71 & 14.52 & 20.17 & 21.10 & 14.20 & 13.39 & 22.13 & 20.37 & 21.83 & 26.23 & 19.00 & 23.99 & 20.88 & 13.77 & 18.09 & 18.83 \\
VICT & 17.57 & 15.21 & 21.01 & 21.93 & 14.56 & 18.29 & 23.35 & 21.38 & 23.00 & 26.42 & 19.39 & 24.56 & 21.86 & 19.58 & 18.46 & 20.44 \\ \hline\midrule
\multicolumn{17}{c}{(f) low-light enhancement LoL-C (PSNR $\uparrow$)} \\ \midrule
\multicolumn{17}{l}{\emph{zero-shot setting:}} \\
Painter & 16.69 & 13.99 & 19.42 & 20.09 & 11.27 & 12.26 & 12.13 & 19.61 & 12.91 & 17.44 & 18.30 & 21.36 & 17.85 & 12.53 & 17.04 & 16.19 \\
VICT & 16.79 & 14.00 & 19.61 & 20.07 & 11.32 & 12.22 & 12.02 & 19.65 & 12.93 & 17.55 & 18.38 & 21.43 & 19.10 & 12.79 & 17.12 & 16.33 \\ \midrule
\multicolumn{17}{l}{\emph{one-shot setting:}} \\
\multicolumn{1}{l}{Painter} & 17.01 & 14.11 & 19.43 & 20.08 & 11.30 & 12.46 & 12.67 & 19.63 & 13.77 & 17.40 & 18.40 & 21.37 & 17.35 & 11.90 & 17.02 & 16.26 \\
VICT & 17.82 & 16.05 & 19.47 & 20.14 & 11.38 & 12.44 & 12.81 & 19.54 & 15.32 & 17.59 & 18.59 & 21.26 & 19.87 & 12.86 & 17.20 & 16.82 \\ \bottomrule
\end{tabular}
\end{table*}

\end{document}